\documentclass[10pt,twocolumn,letterpaper]{article}

\usepackage{eso-pic}
\usepackage{xspace}
\usepackage{times}
\usepackage{epsfig}
\usepackage{float}
\usepackage{graphicx}
\usepackage{amsmath}
\usepackage{amssymb}
\usepackage{algorithm}
\usepackage{algpseudocode}
\usepackage{subfiles}
\usepackage{caption}
\usepackage{subcaption}
\usepackage[utf8]{inputenc}
\usepackage[english]{babel}
\newtheorem{theorem}{Theorem}[section]

\newtheorem{lemma}[theorem]{Lemma}

\begin{document}

\title{Illegible Text to Readable Text: An Image-to-Image Transformation using Conditional Sliced Wasserstein Adversarial Networks}

\author{Mostafa Karimi  ~\thanks{ Work done during an internship at data science team,  Ancestry.com}\\
Texas A$\&$M University\\
{\tt\small mostafa$\_$karimi@tamu.edu}
\and
Gopalkrishna Veni \\
Ancestry.com\\
{\tt\small gveni@ancestry.com}
\and
Yen-Yun Yu \\
Ancestry.com\\
{\tt\small yyu@ancestry.com}
}

\maketitle

\begin{abstract}
Automatic text recognition from ancient handwritten record images is an important problem in the genealogy domain. However, critical challenges such as varying noise conditions, vanishing texts, and variations in handwriting makes the recognition task difficult. We tackle this problem by developing a handwritten-to-machine-print conditional Generative Adversarial network (HW2MP-GAN) model that formulates handwritten recognition as a text-Image-to-text-Image translation problem where a given image, typically in an illegible form, is converted into another image, close to its machine-print form. The proposed model consists of three-components including a generator, and word-level and character-level discriminators. The model incorporates Sliced Wasserstein distance (SWD) and U-Net architectures in HW2MP-GAN for better quality image-to-image transformation. Our experiments reveal that HW2MP-GAN outperforms state-of-the-art baseline cGAN models by almost 30 in Frechet Handwritten Distance (FHD), 0.6 in average Levenshtein distance and 39\% in word accuracy for image-to-image translation on IAM database. Further, HW2MP-GAN improves handwritten recognition word accuracy by 1.3\% compared to baseline handwritten recognition models on IAM database. 
\end{abstract}

\section{Introduction}
\documentclass[main.tex]{subfiles}
Text recognition from ancient handwritten record images is an important problem in the genealogy domain helping genealogists discover and unlock family history. Automating the text recognition process would further benefit them in saving time, manual labor and associated cost respectively. However, ancient document images suffer from critical challenges including varying noise conditions, interfering annotations, typical ancient record artifacts like fading and vanishing texts, and variations in handwriting making it difficult to transcribe~\cite{murdock2015}. Over the past decade, various approaches have been proposed to solve document analysis and recognition such as optical character recognition (OCR)~\cite{mori1999optical}, layout analysis~\cite{o1993document}, text segmentation~\cite{koshorek2018text} and handwriting recognition~\cite{scheidl2018word,graves2009offline,graves2008novel,jo2019handwritten}. Although OCR models have been very successful in recognizing machine print text, they stumble upon handwriting recognition due to aforementioned challenges and connecting characters in the text as compared to machine print ones where the characters are easily separable.  

Unlike standard techniques that transcribe handwriting images by treating them as either a classification or segmentation problem~\cite{xu1992methods,lecun1990handwritten,zhang2007stroke,chen1993variable}, depending upon the context, we follow a different approach. In essence, we formulate handwriting recognition as a text-Image-to-text-Image translation problem where a given image, typically in an illegible form, is transformed into an another image, closer to machine-print, which can then be easily transcribed using OCR-like techniques. By doing so, high-quality results can be achieved even on extremely challenging handwriting images. 

Generative adversarial network (GAN)-based deep generative models have shown a great success in image-to-image translation tasks~\cite{goodfellow2014generative, isola2017image,mirza2014conditional}. Basically, GAN~\cite{goodfellow2014generative} consists of a generator network that tries to map latent space (noise) to the true data distribution while generating fake samples resembling the real ones and a discriminator network that tries to distinguish true samples from the fake ones. Both networks compete against each other until they reach equilibrium. However, GAN inherently suffers from major challenges including non-convergence, mode collapse and a vanishing gradient problem~\cite{arjovsky2017principled}. A variant of GAN called sliced Wasserstein GAN (WGAN)~\cite{wu2019sliced} has been introduced to address these challenges. We use a modified version of sliced WGAN into our framework to translate handwritten text images. In the proposed model, we use a U-Net architecture~\cite{ronneberger2015u} inside the generator as it captures low-level as well as abstract features. For the discriminator part, the proposed model accounts for both word- and character-level errors and underlying high-dimensional distributions leveraged by Wasserstein distance with slice sampling to transcribe a given text.

The key contributions of the proposed framework include:
\begin{itemize}
    \item Develop a novel GAN model with three components including one generator and two discriminators. The model generator tries to fool both discriminators to generate realistic "fake" machine print images with realistic characters. The first discriminator, namely a word-level discriminator, tries to distinguish between "fake" machine print generated images and real ones given a handwriting image. The second discriminator is a character level discriminator that tries to distinguish between "fake" character generation and real ones. 
    \item Develop conditional sliced Wasserstein GAN (cSWGAN) model with Lipschitz continuity constraint as a gradient penalty to convert handwritten images into machine print ones.
    \item Utilize U-Net architecture \cite{ronneberger2015u} inside our model generator, similar to the pix2pix model, for good quality image generation.
\end{itemize}

The rest of the paper is organized as follows. Section 2 reviews the related work. In Section 3, our novel cSWGAN with word- and character-level discriminators is described. Experiments and results are discussed in Section 4.  and final conclusions are offered in Section 5.


\section{Related works}
\documentclass[main.tex]{subfiles}
Handwriting image recognition is traditionally divided into two groups including online \cite{plamondon2000online} and offline recognition  \cite{tappert1990state}. In the online case, the time series of coordinates representing the movement of the pen tip is captured \cite{graves2008novel} whereas in offline, the image of the text is available. We deal with the latter case. Several computer vision and machine learning algorithms have been proposed to solve various challenges of handwriting recognition~\cite{blumenstein2002new,kumar2013analytical} but the problem is far from being solved. Some standard handwriting recognition approaches include hidden Markov models \cite{starner1994line}, support vector machines \cite{bahlmann2002online} and sequential networks including recurrent neural networks (RNN) and its variants. 

Long short term memory (LSTM) networks are a type of RNN that propagate sequential information for long periods of time and have been widely applicable in handwriting recognition tasks~\cite{graves2008novel}. Multidimensional Recurrent Neural Networks \cite{graves2007multi} are another type of sequential networks that have been widely used in modern handwritten text recognition tasks~\cite{graves2009offline}. Annotating handwritten text at a character level is a challenging task. Connectionist Temporal Classification (CTC)~\cite{graves2006connectionist} has been developed that avoids calculating loss of sequential networks at the character level. Further, CTC-based networks do not require post-processing of the recognized text. Therefore sequential networks with CTC loss has gained a lot of attention in handwriting recognition tasks. The proposed design also uses a similar model as a part of its framework.  

As mentioned earlier, Generative adversarial networks (GANs) have proven to be successful generative models in many computer vision tasks. GAN formulates a generative model as a game theory minimax game between generator and discriminator models. The Generator model tries to generate "fake" samples as close to the real ones  and the discriminator model tries to discriminate "fake" samples from real ones. An extension of GAN is conditional GAN where the sample generation is conditioned upon an input which can be a discrete label \cite{mirza2014conditional}, a text \cite{reed2016generative} or an image \cite{isola2017image}. Isola et al.,~\cite{isola2017image} proposed pix2pix GAN that utilizes conditional GAN framework and U-Net architecture ~\cite{ronneberger2015u} for their generator and discriminator models. This approach tends to capture hierarchical features inside images. Although GAN models are very successful in generating fascinating, realistic images~\cite{karras2017progressive}, they are hard to train due to their difficulty in achieving Nash equilibrium \cite{salimans2016improved}, low dimensional support~\cite{arjovsky2017wasserstein}, vanishing gradient~\cite{mao2017least}, and mode collapsing \cite{arjovsky2017wasserstein} issues. 

Existing GANs employ either Kullback–Leibler (KL) or Jensen–Shannon (JS) divergence to model loss functions, which could give rise to mode collapsing, gradient vanishing and low dimensional support problems in a high-dimensional space. Wasserstein distance (WD) has gained attention in computer vision and machine learning community due to its everywhere continuous and almost everywhere differentiable nature, which can overcome the above mentioned problems especially the low dimensional support problem. Arjovsky et al.,~\cite{arjovsky2017wasserstein} proposed Wasserstein GAN (WGAN), which uses Wasserstein-1 (earth mover) distance to learn probability distributions. 
The underlying problem with Wasserstein-1 distance is that its primal form is intractable~\cite{arjovsky2017wasserstein} and it is hard to enforce Lipschitz continuity constraint in high-dimensional space for its dual form. To circumvent this problem, sliced Wasserstein Distance (SWD)~\cite{wu2019sliced} can be used based on the fact that Wasserstein distance provides a closed-form solution for one-dimensional probability densities. 
Previously, SWD has been utilized for dimensionality reduction, clustering~\cite{slicedkolouri2016}, and learning Gaussian mixture models~\cite{kolouri2018sliced}. Recently, it has been employed in generative models such as sliced Wasserstein generative models \cite{wu2019sliced} and sliced Wasserstein auto-encoders~\cite{kolouri2018sliced}. 
SWD factorizes high-dimensional probabilities to multiple marginal distributions~\cite{wu2019sliced}. Theoretically, SWD can compute infinitely many linear projections of a high-dimensional distribution to one-dimensional distributions followed by computing average Wasserstein distance of these one-dimensional distributions \cite{kolouri2019generalized}. 

We have developed a novel conditional sliced Wasserstein GAN with three components including a generator, a word-level discriminator and a character-level discriminator for translating handwritten text images to corresponding machine print forms.

\section{Methods}
\documentclass[main.tex]{subfiles}
\subsection{Generative adversarial networks}
Formally, the objective function of GAN is written as:

\begin{equation}
    \label{eq:GAN}
    \min_{G}\max_{D} \underset{\mathbf{x} \sim \mathbb{P}_r}{\mathbb{E}}[\log{(D(\mathbf{x}))}] + \underset{\tilde{\mathbf{x}} \sim \mathbb{P}_g}{\mathbb{E}}[\log{(1-D(\tilde{\mathbf{x}}))}],
\end{equation}
where $G$ represents a generator, $D$ represents a discriminator and $\mathbf{x}$ is the realization of true samples. $\mathbb{P}_r$ is the true data distribution and $\mathbb{P}_g$
denotes the generator's distribution that is modeled implicitly by $\tilde{\mathbf{x}} \sim G(\mathbf{z})$ and $\mathbf{z} \sim \mathbb{P}(\mathbf{z})$ (the latent space or noise $\mathbf{z}$ is sampled usually from a uniform distribution or a spherical Gaussian distribution).

Training a GAN network is equivalent to minimizing the Jensen-Shannon (JS) divergence between $\mathbb{P}_r$ and $\mathbb{P}_g$ if the discriminator is trained to optimality before each generator's update \cite{goodfellow2014generative}. However, it has been observed that Eq.~(\ref{eq:GAN}) tends to suffer from the gradient vanishing problem as the discriminator saturates. Although generator's loss function can be replaced by maximizing $\underset{\mathbf{z} \sim \mathbb{P}(\mathbf{z})}{\mathbb{E}}[\log{(D(G(\mathbf{z})))}]$, the gradient vanishing is far from being solved~\cite{goodfellow2014generative}.

Later, GAN has been extended to conditional GAN (cGAN) \cite{mirza2014conditional} where both generator and discriminator are conditioned on a given additional supervised event $\mathbf{y}$, where $\mathbf{y}$ can be any kind of auxiliary information or data such as discrete label \cite{mirza2014conditional}, text \cite{reed2016generative} and image \cite{isola2017image}. Usually cGAN is performed by feeding $\mathbf{y}$ into both discriminator and generator as an additional input layer. cGAN is formulated as:
\begin{equation}
    \label{eq:CGAN}
    \min_{G}\max_{D} \underset{\mathbf{x} \sim \mathbb{P}_r}{\mathbb{E}}[\log{(D(\mathbf{x}|\mathbf{y}))}] + \underset{\tilde{\mathbf{x}} \sim \mathbb{P}_g}{\mathbb{E}}[\log{(1-D(\tilde{\mathbf{x}}|\mathbf{y}))}]
\end{equation}
where $\mathbb{P}_g$, the generator's distribution, is explicitly modeled as $\tilde{\mathbf{x}} \sim G(\mathbf{z}|\mathbf{y})$ and $\mathbf{z} \sim \mathbb{P}(\mathbf{z})$ in cGAN.

\subsection{Wasserstein distance}
Wasserstein distance is a powerful metric in the field of optimal transport and has recently drawn a lot of attention~\cite{kolouri2018sliced}. It measures the distance between two distributions.
p-Wasserstein distance (WD) between two random variables X, Y is given as:
\begin{equation}
 W_p = \underset{\gamma \in \Gamma(\mathbb{P}_X,\mathbb{P}_Y)}{\inf}  \,\,\, \underset{(\mathbf{x},\mathbf{y})\sim \gamma}{\mathbb{E}}[d^{p}(\mathbf{x},\mathbf{y})]^{\frac{1}{p}},
\end{equation}
where $\Gamma(\mathbb{P}_X,\mathbb{P}_Y)$ denotes a set of all joint distributions $\gamma(X,Y)$ whose marginal distributions are $\mathbb{P}_X$, $\mathbb{P}_Y$. Suppose $x$ and $y$ are realizations or samples from random variables $X$ and $Y$ respectively. 
Let $p > 0$, then $d(\mathbf{x},\mathbf{y})$ defines a metric for $\mathbf{x}$ and $\mathbf{y}$.
For $p = 1$, then 1-WD $d(\mathbf{x},\mathbf{y})$ is named as Earth-Mover distance (EMD). Intuitively, $\gamma(X,Y)$ shows how much "mass" is going to be transported from any realization of $\mathbf{X}$ to any realization of $\mathbf{Y}$ in order to transport distribution $\mathbb{P}_X$ to the distribution $\mathbb{P}_Y$. Because the primal form of the 1-WD is generally intractable and usually the dual form is used in practice\cite{arjovsky2017wasserstein},
a dual form of EMD is formulated through the Kantorovich-Rubinstein (KR) duality \cite{arjovsky2017wasserstein}
and is given as:
\begin{equation}
   \label{eq:dual}
   W_1 = \underset{||g||_L \le 1}{\sup} \,\,\, \underset{\mathbf{x} \sim \mathbb{P}_X}{\mathbb{E}}[g(\mathbf{x})]-\underset{\mathbf{y} \sim \mathbb{P}_Y}{\mathbb{E}}[g(\mathbf{y})],
\end{equation}
where the supremum is over all the 1-Lipschitz functions $g(\cdot)$. 

\subsection{Wasserstein Generative adversarial networks}
One challenge in applying WD on GAN is that WD is a much weaker distance compared to the JS distance, i.e., it induces a weaker topology. This fact makes a sequence of probability distributions converge in the distribution space~\cite{arjovsky2017wasserstein}, which results in bringing the model distribution closer to the real distribution. 
In other words,
both the low dimensional support challenge in high dimensions and the gradient vanishing problem could be solved under this assumption. 
Due to all of these reasons, the Wasserstein GAN (WGAN) model has been developed based on the dual form of the EMD \cite{arjovsky2017wasserstein}.  WGAN with generator $G$ and discriminator $D$ is formulated as the first term of  Eq.~(\ref{eq:WGANGP}).
The main challenge in WGAN is to satisfy the Lipschitz continuity constraint. The original WGAN considered a weighted clipping approach that limits the capacity of the model and its performance ~\cite{gulrajani2017improved}. To alleviate this problem, WGAN with gradient penalty (WGAN-GP) \cite{gulrajani2017improved} has been developed that penalizes the norm of the discriminator’s gradient with respect to a few input samples. The gradient penalty $GP = \underset{\hat{\mathbf{x}} \sim \mathbb{P}_{\hat{X}}}{\mathbb{E}}[(||\nabla_{\hat{\mathbf{x}}} D(\hat{\mathbf{x}})||_2 - 1)^2]$ is added to the original WGAN loss function in Eq.~(\ref{eq:WGANGP}). Therefore, WGAN-GP is formulated as:
\begin{equation}
   \begin{split}
    \label{eq:WGANGP}
    \underset{G}{\min} \, \underset{||D||_L \le 1}{\max} \,\,\, 
    \underset{\mathbf{x} \sim \mathbb{P}_r}{\mathbb{E}}[D(\mathbf{x})]
    &
    -\underset{\mathbf{y} \sim \mathbb{P}_g}{\mathbb{E}}[D(\mathbf{y})] + \lambda GP
    \end{split}
\end{equation}

$\hat{\mathbf{x}}$ represents random samples following the distribution $\mathbb{P}_{\hat{X}}$,
which is formed by uniformly sampling along the straight lines between pair of points sampled from $\mathbb{P}_r$ and $\mathbb{P}_g$.
$\lambda$ is the hyper-parameter to balance between original WGAN loss function and the gradient penalty regularization. Recently WGAN has been further improved by adding consistency term GAN (CTGAN) \cite{wei2018improving}.

\subsection{Sliced wasserstein distance (SWD)}
WD is generally intractable  for multi-dimensional probability distribution~\cite{kolouri2019generalized}. However, there is a closed-form solution (i.e., WD is tractable) if the distribution is in the low-dimensional space (in this paper, we use an one dimensional space). Let $F_X$ and $F_Y$ be the cumulative distribution function (CDF) for probability distributions $\mathbb{P}_X$ and $\mathbb{P}_Y$ respectively. The WD between these two distributions is uniquely defined as $F_Y^{-1}(F_X(x))$.The primal p-WD between them can be re-defined as:
\begin{equation}
      W_{p}  = \Bigg(\int_{0}^{1} d^p(F_X^{-1}(z),F_Y^{-1}(z)) dz\Bigg)^{\frac{1}{p}}
    \label{WD_1}
\end{equation}
The change of variable $z:= F_x(x)$ is used to derive the equation. For empirical distributions, Eq.~(\ref{WD_1}) is calculated by sorting two distributions and then calculating the average distance $d^p(\cdot,\cdot)$ between two sorted samples which requires $O(M)$ at best and $O(M \log M)$ at worst, where $M$ is number of samples for each distribution\cite{kolouri2019generalized}. 

Sliced Wasserstein distance (SWD) utilizes this property by factorizing high-dimensional probabilities to multiple marginal distributions \cite{wu2019sliced} with standard Radon transform, denoted by $\mathcal{R}$. Given any distribution $P(\cdot)$, the Radon transform of $P(\cdot)$ is defined as:
\begin{equation}
    \mathcal{R}P(t,\theta) = \int_{\mathbb{R}^d} \mathbb{P}(\mathbf{x}) \delta(t-\langle \mathbf{\theta},\mathbf{x}\rangle) d\mathbf{x},
\end{equation}
where $\delta(\cdot)$ is the one-dimensional Dirac delta function and $\langle \cdot, \cdot \rangle$ is the Euclidean inner-product. 
The hyper-parameters in Radon transform include level set parameter $t \in \mathbb{R}$ and normal vector $\theta \in \mathbb{S}^{d-1}$ ($\theta$ is a unit vector, and  $\mathbb{S}^{d-1}$ is the unit hyper-sphere in d-dimensional space).  
Radon transform $\mathcal{R}$ maps a function to the infinite set of its integrals over the hyperplanes $\langle \mathbf{\theta},\mathbf{x}\rangle$ of $\mathbb{R}^d$. For a fixed $\theta$, the integrals over all hyperplanes define a continuous function $\mathcal{R}P(.,\theta): \mathcal{R}\rightarrow \mathcal{R}$ which is a slice or projection of $P$. 
The p-WD in Eq.~(\ref{WD_1}) can be rewritten as the sliced p-WD for a pair of distributions $\mathbb{P}_X$ and $\mathbb{P}_Y$:
\begin{equation}
    \label{eq:SWGAN_ori}
    SW_p = \Bigg(\int_{\mathbb{S}^{d-1}} W_p(\mathcal{R}P_X(.,\theta),\mathcal{R}P_Y(.,\theta)) d\theta \Bigg)^{\frac{1}{p}}
\end{equation}
The dual of Eq.~(\ref{eq:SWGAN_ori}) can be derived based on KR duality:
\begin{equation}
    SW_p = \Bigg(\int_{\mathbb{S}^{d-1}} \underset{||g||_L \le 1}{\sup} \,\underset{\mathbf{x_{\theta}}}{\mathbb{E}}[g(\mathbf{x_{\theta}})] -\underset{\mathbf{y_{\theta}} }{\mathbb{E}}[g(\mathbf{y_{\theta}})] \Bigg)^{\frac{1}{p}}
    \label{eq:SWD_dual}
\end{equation}
where $x_\theta$ and $y_\theta$ are sampled from $\mathcal{R}P_X(.,\theta)$ and $\mathcal{R}P_Y(.,\theta)$ respectively. SWD is not only a valid distance which satisfies positive- definiteness, symmetry and the triangle inequality \cite{wu2019sliced}, but also equivalent to WD based on Lemma~\ref{lemma_1}.
\begin{lemma}
Following inequality holds for SWD and WD where $\alpha_1$ and $\alpha_2$ are constants and n is the dimension of sample vectors from X and Y \cite{wu2019sliced}:
\begin{equation*}
    SW_p(\mathbb{P}_X,\mathbb{P}_Y)^p \le \alpha_1 W_p(\mathbb{P}_X,\mathbb{P}_Y)^p \le \alpha_2 SW_p(\mathbb{P}_X,\mathbb{P}_Y)^{\frac{1}{n+1}}
    \label{lemma_1}
\end{equation*}
\end{lemma}
    \label{ineq:SWD}

\subsection{Sliced wasserstein Generative adversarial networks (SWGAN)}
Recently Sliced Wasserstein Generative adversarial network (SWGAN) \cite{wu2019sliced} has been proposed based on utilizing and approximating the SWD in generative models. Motivated from the WGAN, \cite{wu2019sliced} has developed their model based on the dual SWD. Their discriminator is the composition of the encoding network $E$ combined with M dual SWD blocks $\{S_{m}\}_{m=1}^M$, that is, $D := \{S_{m} \circ E\}_{m=1}^{M}$. The encoder $E : \mathcal{R}^{b \times n} \rightarrow \mathcal{R}^{b \times r}$ will map the batch of data $X \in \mathcal{R}^{b \times n}$ to the latent space of $X^\text{embd} \in \mathcal{R}^{b \times r}$ where $b$ is the batch size, $n$ is the data dimension and $r$ is the latent dimension. Then, the first part of each dual SWD block will operate an orthogonalization operation $X^\text{orth}= X^\text{embd} \Theta$ with $\Theta \in \mathcal{R}^{r \times r}$ to make sure the encoded matrix is orthogonal. The second part of each dual SWD block will perform an element-wise non-linear neural network function $T_{i}(\mathbf{x}_i^\text{orth}) = u_i \text{LeakyReLU}(w_i \mathbf{x}_i^\text{orth} + b_i)$ to approximate the one-dimensional optimal $g$ function \cite{wu2019sliced} in Eq. (\ref{eq:SWD_dual}) for all $i = 1, ..., r$ where $u_i, w_i ,\text{and}    \, b_i$ are scalar parameters. 
Eventually, the model can be approximated by taking in integral over $\mathbb{S}^{n-1}$ and summing over the output mean value of the dual SWD blocks. 

The Lipschitz constraint can be easily applied for one-dimensional functions followed by the gradient penalty on each dimension of the $T_i$'s. During learning, the projection matrices should remain orthogonal. In order to keep projection matrices orthogonal throughout the training process, a manifold-valued update rule has been developed based on the Stiefel manifolds \cite{wu2019sliced}.
SWGAN's final objective function is follows:
\begin{equation}
   \begin{split}
    &\underset{G}{\min} \, \underset{D}{\max} \int_{\theta \in \mathbb{S}^{n-1}} \,\,\underset{\mathbf{x} \sim \mathbb{P}_r}{\mathbb{E}}[D(\mathbf{x})]-\underset{\mathbf{y} \sim \mathbb{P}_g}{\mathbb{E}}[D(\mathbf{y})]+\\
    & \lambda_1 \underset{\hat{\mathbf{x}} \sim \mathbb{P}_{\hat{X}}}{\mathbb{E}}[||\nabla_{\hat{\mathbf{x}}} D(\hat{\mathbf{x}})||_2^2] + \lambda_2 \underset{\hat{\mathbf{y}} \sim \mathbb{P}_{\hat{y}}}{\mathbb{E}}[(||\nabla_{\hat{\mathbf{y}}} T(\hat{\mathbf{y}}) - \mathbf{1} ||_2^2]
    \end{split}
\end{equation}
where $\theta$ represents trainable parameters embedded in $D$, $\mathbf{1}$ is a vector with all entries equal to 1, $\lambda_1$ and $\lambda_2$ are the hyper-parameters for balancing the gradient penalty terms and the dual SWD.

\subsection{Proposed method}
In this paper, we have developed a handwritten-to-machine print GAN (HW2MP-GAN) model to pre-process and convert handwritten text images to machine print ones. We consider a \textit{three-component game} between a single generator and two discriminators, which are character- and word-level discriminators, for our conditional GAN model. 
Two discriminators work together and help the generator in producing clear words and characters in the correct order. 
The \textit{character}-level discriminator enforces each generated character to be similar to real machine-print characters. Since the number of English characters, symbols and numbers is limited, the \textit{character}-level discriminator's task of learning to generate each one of these characters correctly is easier than the other one. The \textit{word}-level discriminator forces generated words to be similar to the real ones. Since the number of combination of all characters, symbols and numbers is exponential to the length of the word, \textit{word}-level discriminator does a harder task of enforcing the correct order from the generated characters. 

\textbf{Character level}:
Suppose, real and generated machine print images are $\mathbf{x}$ and $\mathbf{\tilde{x}}$ respectively. Assume there are $K_{\mathbf{x}}$ characters in the image $\mathbf{x}$. Then, we define the real and generated machine print characters as $\{\mathbf{x}^c_k\}_{k =1}^{K_{\mathbf{x}}}$ and $\{\mathbf{\tilde{x}}^c_k\}_{k =1}^{K_{\mathbf{x}}}$ respectively. Superscript "c" and "w" are used for character-level and word-level respectively. $\mathbf{x}^c_k$ and $\mathbf{\tilde{x}}^c_k$ represent the $k^{th}$ character of word $\mathbf{x}$ and $\mathbf{\tilde{x}}$ respectively. Our character level discriminator is defined as $D^c := \{S_{m}^c \circ E^c\}_{m=1}^{M^c}$ where $E^c$ is the character level encoder, $S_{m}^c$ is the $m^{th}$ SWD block and $M^c$ is the number of SWD blocks for character-level discriminator. Therefore, the character-level loss function is formulated as:
\begin{equation}
    \begin{split}
        &L^c = \int_{\mathbf{\theta}^c \in \mathbb{S}^{r^c-1}} \,\,\,\underset{\mathbf{x}^c_k \sim \mathbb{P}_r^c}{\mathbb{E}}[D^c(\mathbf{x}^c_k)]-\underset{\mathbf{\tilde{x}}^c_k \sim \mathbb{P}_g^c}{\mathbb{E}}[D^c(\mathbf{\tilde{x}}^c_k)]+\\
    & \lambda_1^c \underset{\mathbf{\hat{x}}^c_k \sim \mathbb{P}_{\mathbf{\hat{x}}}^c}{\mathbb{E}}[||\nabla_{\mathbf{\hat{x}}^c_k} D^c(\mathbf{\hat{x}}^c_k)||_2^2] +\lambda_2^c \underset{\mathbf{\bar{x}}^c_k \sim \mathbb{P}_{\mathbf{\bar{x}}}^c}{\mathbb{E}}[(||\nabla_{\mathbf{\bar{x}}^c_k} T^c(\mathbf{\bar{x}}^c_k) - \mathbf{1} ||_2^2]
    \end{split}
    \label{eq:lc}
\end{equation}

where the real machine print character distribution is $\mathbb{P}_r^c$ and the generated machine print character distribution is $\mathbb{P}_g^c$. $\mathbf{\theta}^c$ represent learnable parameters and are embedded in the character discriminator $D^c$. 
The last two terms of Eq.~(\ref{eq:lc}) are gradient and Lipschitz regularization terms, where hyper-parameters $\lambda_1^c$ and $\lambda_2^c$ are balancing between the SWGAN's loss function and its regularization terms, and $\mathbf{1}$ is the vector of all ones. 
The gradient and Lipschitz regularization are enforced according to the $\mathbb{P}_{\mathbf{\hat{x}}}^c$ and $\mathbb{P}_{\mathbf{\bar{x}}}^c$ distributions which are sampling across the lines between $\mathbb{P}_r^c$ and $\mathbb{P}_g^c$.

\textbf{Word level}:
Similarly to character-level discriminator, our word-level discriminator is defined as 
$D^w := \{S_{m}^w \circ E^w\}_{m=1}^{M^w}$ where the $E^w$ is word level encoder, $S_{m}^w$ is $m^{th}$ SWD block and $M^w$ is the number of SWD blocks. Therefore, the word level loss function is formulated as:
\begin{equation}
   \begin{split}
    &L^w = \int_{\theta^w \in \mathbb{S}^{n-1}} \,\,\,\underset{\mathbf{x} \sim \mathbb{P}_r}{\mathbb{E}}[D^w(\mathbf{x}|\mathbf{y})]-\underset{\mathbf{\tilde{x}} \sim \mathbb{P}_g}{\mathbb{E}}[D^w(\mathbf{\tilde{x}}|\mathbf{y})] \\
    &+ \lambda_1^w \underset{\hat{\mathbf{x}} \sim \mathbb{P}_{\hat{x}}}{\mathbb{E}}[||\nabla_{\hat{\mathbf{x}}} D^w(\hat{\mathbf{x}}|\mathbf{y})||_2^2] + \lambda_2^w \underset{\mathbf{\bar{x}} \sim \mathbb{P}_{\bar{x}}}{\mathbb{E}}[(||\nabla_{\mathbf{\bar{x}}} T^w(\mathbf{\bar{x}}) - \mathbf{1} ||_2^2] \Big]
    \end{split}
    \label{eq:lw}
\end{equation}
where the real machine print word distribution is $\mathbb{P}_r$ and the generated machine print word distribution is $\mathbb{P}_g$. $\mathbf{\theta}^w$ is the learnable parameters and embedded in the word discriminator $D^w$. The last two terms are the gradient and Lipschitz regularization terms where hyper-parameters $\lambda_1^w$ and $\lambda_2^w$ are balancing between the SWGAN's loss function and its regularization terms. Similarly, the gradient and Lipschitz regularization are enforced according to the $\mathbb{P}_{\mathbf{\hat{x}}}$ and $\mathbb{P}_{\mathbf{\bar{x}}}$ distributions.



\textbf{Proposed HW2MP-GAN:}
Our final loss function is combined with character level model, Eq.~(\ref{eq:lc}), and word level model, Eq.~(\ref{eq:lw}), with reconstruction loss, which is the $l_1$ norm between generated images $\tilde{x}$ and real images $x$. The objective function of HW2MP-GAN is:
\begin{equation}
    L^{total} = L^w + \lambda_{char} L^c + \lambda_{recons}
    \underset{\substack{\mathbf{x} \sim \mathbb{P}_r\\\mathbf{\tilde{x}} \sim \mathbb{P}_g}}{\mathbb{E}} ||\mathbf{\tilde{x}} - \mathbf{x}||_1
    \label{eq:lwhole}
\end{equation}
where $\lambda_{char}$ and $\lambda_{recons}$ are hyper-parameters for balancing between word-level loss, character-level loss and the reconstruction loss functions. To make sure that the projection matrices are orthogonal during training for both character- and word-level discriminators, we follow the Steifel manifold similar to \cite{wu2019sliced}. 

Our whole pipeline is illustrated in figure~\ref{fig:whole_pipeline} (a) and the pseudo-code of our algorithm is written in algorithm (\ref{alg:HW2MP}).
\begin{figure*}
    \centering
    \includegraphics[width=0.7\textwidth]{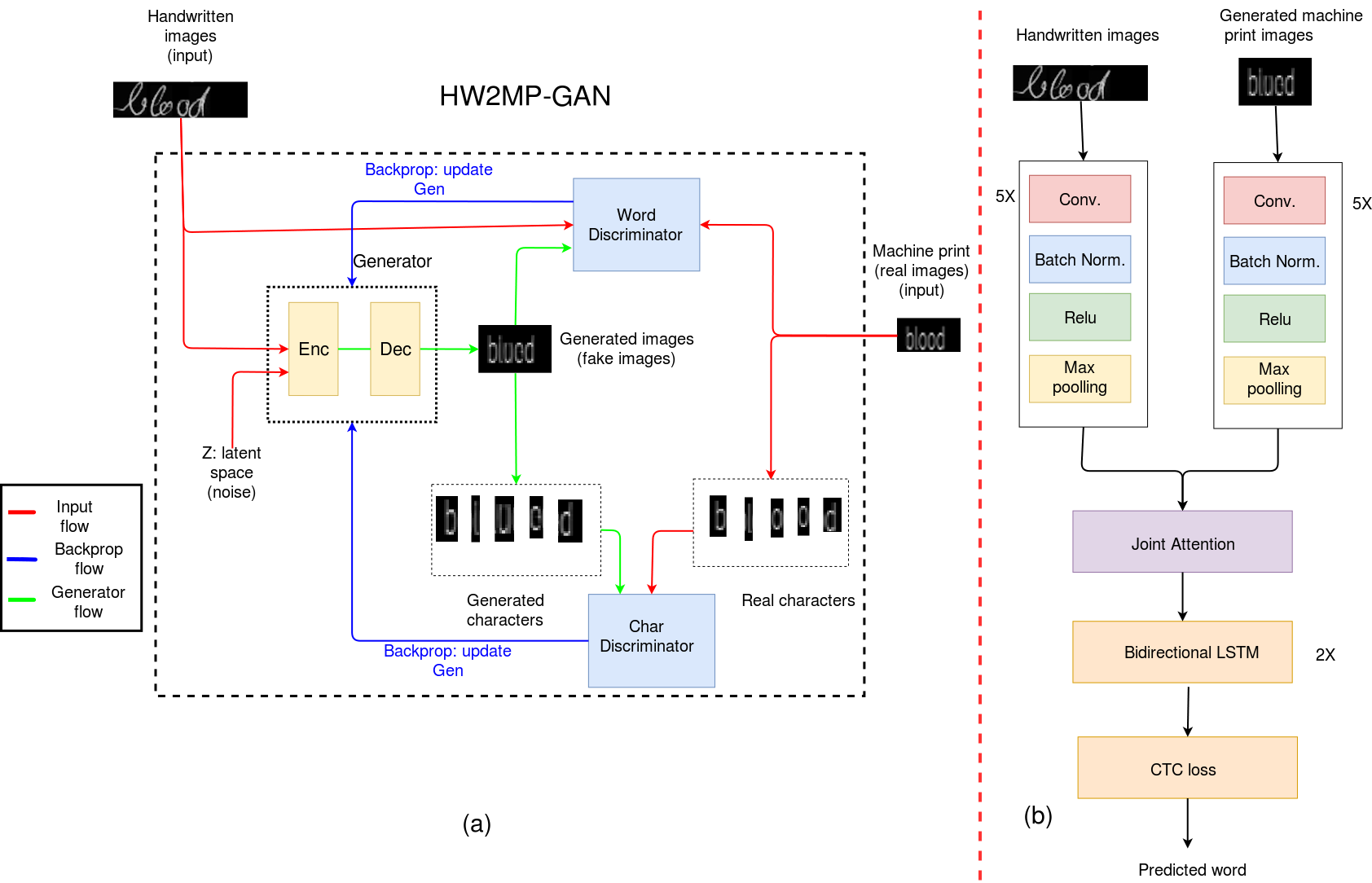}
    \caption{(a) Overall pipeline for HW2MP-GAN (b) Joint attention handwriting recognition reinforced by HW2MP-GAN}
    \label{fig:whole_pipeline}
\end{figure*}
\begin{algorithm}
\caption{HW2MP-GAN}\label{alg:HW2MP}
\begin{algorithmic}[1]
\footnotesize
\Require Number of dual SWD blocks for word and character level discriminators are $M^w$ and $M^c$, batch size b, generator G, word level discriminator $D^w = [S_{d,1}^w \circ E^w,\cdots,S_{d,M^w}^w \circ E^w]^T$ and character level discriminator $D^c = [S_{d,1}^c \circ E^c,\cdots,S_{d,M^c}^c \circ E^c]^T$, latent code dimension for word and character level discriminators are $r^w$ and $r^c$, Lipschitz constants $k^c$ and $k^w$, training steps h, training hyper-parameters,etc.
\For{iter=$ 1 \cdots n_{max}$}
\For{t=1$ \cdots n_{critic}$}
    \State Sample real data $\{\mathbf{x}^{(i)},\mathbf{y}^{(i)}\}_{i=1}^{m} \sim \mathbb{P}_r$
    \State Sample noise $\{\mathbf{z}^{(i)}\}_{i=1}^{m} \sim p(\mathbf{z})$
    \State Sample random number $\{\epsilon_1^{(i)}\}_{i=1}^{m},\{\epsilon_2^{(i)}\}_{i=1}^{m} \sim U[0,1]$
    \State $\{\mathbf{\tilde{x}}^{(i)}\}_{i=1}^{m} \leftarrow \{G_{\theta}(\mathbf{z}^{(i)}|\mathbf{y}^{(i)})\}_{i=1}^{m}$
    \State $\{\mathbf{\hat{x}}^{(i)}\}_{i=1}^{m} \leftarrow \{\epsilon_1^{(i)} \mathbf{x}^{(i)} + (1-\epsilon_1^{(i)}) \mathbf{\tilde{x}}^{(i)}\}_{i=1}^{m}$
    \State $\{\mathbf{\bar{x}}^{(i)}\}_{i=1}^{m} \leftarrow \{\epsilon_2^{(i)} \mathbf{x}^{(i)} + (1-\epsilon_2^{(i)}) \mathbf{\tilde{x}}^{(i)}\}_{i=1}^{m}$
    \State $L^w$ is defined in Eq. (\ref{eq:lw})
    \State $\theta^w \leftarrow \text{Adam}(\nabla_{\theta^w} \frac{1}{m} \sum_{i=1}^{m} L^w,\theta^w,\alpha,\beta_1,\beta_2)$
\EndFor
\For{t=1$ \cdots n_{critic}$}
    \State Sample real character data $\{\mathbf{x}^{c,(i)}\}_{i=1}^{m} \sim \mathbb{P}_r$
    \State Sample noise $\{\mathbf{z}^{(i)}\}_{i=1}^{m} \sim p(\mathbf{z})$
    \State Sample random number $\{\epsilon_1^{(i)}\}_{i=1}^{m},\{\epsilon_2^{(i)}\}_{i=1}^{m} \sim U[0,1]$
    \State $\{\mathbf{\tilde{x}}^{c,(i)}\}_{i=1}^{m} \leftarrow \{G_{\theta}(\mathbf{z}^{(i)})\}_{i=1}^{m}$
    \State $\{\mathbf{\hat{x}}^{c,(i)}\}_{i=1}^{m} \leftarrow \{\epsilon_1^{(i)} \mathbf{x}^{c,(i)} + (1-\epsilon_1^{(i)}) \mathbf{\tilde{x}}^{c,(i)}\}_{i=1}^{m}$
    \State $\{\mathbf{\bar{x}}^{c,(i)}\}_{i=1}^{m} \leftarrow \{\epsilon_2^{(i)} \mathbf{x}^{c,(i)} + (1-\epsilon_2^{(i)}) \mathbf{\tilde{x}}^{c,(i)}\}_{i=1}^{m}$
    \State $L^c$ is defined in Eq. (\ref{eq:lc})
    \State $\theta^c \leftarrow \text{Adam}(\nabla_{\theta^c} \frac{1}{m} \sum_{i=1}^{m} L^c,\theta^c,\alpha,\beta_1,\beta_2)$
\EndFor
\State Sample a batch of noises $\{\mathbf{z}^{(i)}\}_{i=1}^{m} \sim p(\mathbf{z})$
\State $L^{total}$ is defined in Eq. (\ref{eq:lwhole})
\State $\theta^g \leftarrow \text{Adam}(\nabla_{\theta^g} \frac{1}{m} \sum_{i=1}^{m} L^{total},\theta^g,\alpha,\beta_1,\beta_2)$
\EndFor
\end{algorithmic}
\end{algorithm}

\subsection{Handwriting recognition reinforced by HW2MP-GAN}
\label{method:recognition}
As explained in previous subsections, we have developed a novel conditional GAN model, HW2MP-GAN, for converting handwritten images to machine print ones. We have further developed a novel attention-based handwriting recognition model that exploits both handwritten images and their HW2MP-GAN generated machine print ones for the handwriting recognition task. As a proof-of-concept, we have modified a standard handwriting recognition model developed by Shi et al. \cite{shi2016end} to exploit both handwritten and generated machine print images. The baseline model consists of CNN layers followed by bidirectional LSTM layers followed by a Connectionist Temporal Classification (CTC) loss \cite{graves2006connectionist}. Further, for posterior decoding of CTC loss to predict the words, we used the recently proposed Word beam search algorithm \cite{scheidl2018word}.

We have developed a novel joint attention handwriting recognition model reinforced by HW2MP-GAN as illustrated in figure \ref{fig:whole_pipeline} (b). Our model consists of two parallel series of convolutional layer followed by batch normalization, ReLU nonlinearity and max pooling which is repeated 5 times. These two paths of information have been merged together with a novel joint attention model followed by two layers of  Bidirectional LSTMs and CTC loss. The joint attention layer consists of two inputs: 1) features learned from handwritten images denoted by $H=(H_1,\cdots,H_i,\cdots,H_T) \in \mathcal{R}^{T\times d_1}$, and 2) features learned from generated machine print images denoted by $P=(P_1,\cdots,P_j,\cdots,P_T) \in \mathcal{R}^{T\times d_2}$ where $T$ is the maximum length of the word, and $d_1$ and $d_2$ represent the number of features for handwritten images and generated machine ones respectively. Therefore, the joint attention layer is formulated as:
\begin{equation}
    \begin{split}
    N_{ij} &= tanh(H_i W P_j) \textit{,}\,\, \alpha_{ij} = \frac{exp(N_{ij})}{\sum_k exp(N_{ik})} \,\, \forall\, i,j\\
    \hat{H}_{i} &= \sum_j \alpha_{ij} P_j \,\,\,\,\, \forall i\,\,\, \textit{,}\,\,\, A = Concat(H,\hat{H})
    \end{split}
\end{equation}
where $\alpha_{ij}$ represents the similarity between the $i^{th}$ handwritten image character and the $j^{th}$ generated machine print character. $\hat{H}_i$ is the projection features learned from the generated machine print image to the handwritten one through attention model. Finally, the output of the attention layer denoted by $A \in \mathcal{R}^{T\times(d_1+d_2)}$ is a concatenation of the features of handwritten images and their projected ones. 

\section{Experimental Evaluation}
\documentclass[main.tex]{subfiles}
\subsection{Data}
\label{result:data}
We evaluated HW2MP-GAN and our joint attention handwriting recognition models on the IAM handwritten database \cite{marti2002iam}. The IAM database contains 115,320 isolated and labeled words. We randomly chose 95\% of the data for our training set and the remaining 5\% for our test set. Because IAM images have varying sizes, we resized them to $32\times 128$ pixels. Further, we preprocessed all images by standardizing them to zero-mean and unit-variance.

Training of the HW2MP-GAN model requires handwritten text images and corresponding manually generated machine print forms (i.e., "real" machine print images), which can be created through the ground truth labeled words. 
Since machine print images contain individual characters, they are used to calculate character-level model loss. Because we have created the "real" machine print images manually, the position of each character is known. 
Because the number of characters in words varies, we only extracted real or generated characters and ignored the background by enforcing loss zero for the backgrounds. 

\subsection{Evaluation metrics}
For comprehensive evaluation of our model against the state-of-art generative models, we considered three metrics for 1) image-to-image translation problem and 2) handwriting text recognition task.
%
First, Frechet Inception Distance (FID) is the state-of-the-art metric for evaluating the performance of image-to-image generative models. 
It compares distances between a pair of Inception embedding features from real and generated images~\cite{shmelkov2018good}.
In this paper, we extended the FID score to \textit{Frechet Handwritten Distance} (FHD) to calculate the distance between embedded features of real and model generated text images.
The embedded features are computed from 
the output of bidirectional LSTM layers of the pre-trained handwriting recognition model~\footnote{We pre-trained handwriting recognition model~\cite{shi2016end}~\label{note_1} using manually generated machine print images from Sec.~\ref{result:data}, i.e., built an OCR-like recognition model, whose accuracy $>$ 99\% on machine print text images.}.
FHD=0 is the best and it signifies that the embedded features are identical.
%
For the handwritten text recognition task, this paper used average Levenshtein distance (LD=0 is the best)~\cite{levenshtein1966binary} and word accuracy (100\% is the best).

\subsection{Implementation}
As explained earlier, our HW2MP-GAN consists of three components: generator, character-level discriminator, and word-level discriminator. 
Our generator architecture comprise a U-Net model~\cite{ronneberger2015u} with 5 layers of encoder and decoder each, where encoder and decoder are inter-connected through skip connections. 
The architecture of encoders for both character-level and word-level discriminators are similar to the discriminator in WGAN-GP~\cite{gulrajani2017improved} minus the last linear projection layer.
The character-level and word-level encoders embed images to $r^w=128$ and $r^c=32$ features respectively. Similar to the original SWGAN~\cite{wu2019sliced}, we used $M^c=M^w=4$ SWD blocks for both character-level and word-level discriminator . We chose hyper-parameters based on grid search over a limited set and our results can be further improved by increasing the search space of hyper-parameters. 
We chose $\lambda_{char}=2$, $\lambda_{recons}=100$, $\lambda_1^c = \lambda_1^w = 20$ and $\lambda_2^c = \lambda_2^w = 10$. 
Adam optimizer \cite{kingma2014adam} with initial learning rate of 0.0001 was used for training the generator and two discriminators. 

%
\subsection{Text-Image-to-Text-Image translation problem using HW2MP-GAN}
This section talks about the performance of the proposed HW2MP-GAN for solving the Text-Image-to-Text-Image translation problem. 
The experiments include 
1) measuring the distance between real machine print images and HW2MP-GAN generated text images, and 
2) the legibility of HW2MP-GAN generated text images.
To evaluate the legibility, we used a pretrained handwriting recognition model~~\textsuperscript{\ref{note_1}} to recognize the HW2MP-GAN generated text images.
We compared the HW2MP-GAN model with state-of-the-art GANs that include DCGAN \cite{radford2015unsupervised}, LSGAN \cite{mao2017least}, WGAN \cite{arjovsky2017wasserstein}, WGAN-GP \cite{gulrajani2017improved}, CTGAN \cite{wei2018improving}, SWGAN \cite{wu2019sliced} and Pix2Pix \cite{isola2017image}. 
In order to put these GANs (except Pix2Pix) in a framework of converting handwriting text images to machine print ones, we further extended them to conditional GAN by embedding handwritten images to a latent space and then concatenating them with noise for machine print generation. 

The results of IAM dataset evaluation based on the three metrics including FHD, average LD and word accuracy have been reported in Table \ref{tab:image2image_IAM}.
Based on our results, we can categorize them into four groups. First, DCGAN and LSGAN models didn't converge due to gradient vanishing problem; Second, WGAN and Pix2Pix models were better than category-1 GAN models since they have improved the GAN model through a better distance metric (Wasserstein in comparison to JS) and better architecture (U-Net model) but have the worst performances compared to other three models. Third, WGAN-GP, CTGAN and SWGAN turned out to be the best baseline models which have comparable results among themselves and outperformed other baseline models. These models as explained, they either have better WD approximation (SWGAN) or better enforcing of Lipschitz continuity constraint (WGAN-GP and CTGAN). Fourth, HW2MP-GAN model outperformed others with a large margin by using all the three metrics. The superior performance of HW2MP-GAN is due to the three-component game, exploiting SWD distance, U-Net architecture and L1 reconstruction loss. 
However, none of these factors considering alone led to this improvement since for example U-Net architecture and L1 reconstruction loss exist in Pix2Pix model and the SWD distance exists in SWGAN.

Test examples have been illustrated in Figure~\ref{fig:results}. Based on these results, we can observe that generated machine print images are very similar to the "real" machine print ones. 
Some errors have been noticed in generating machine print images 
for example 1) "d" instead of "o" in the word "Almost" 
2) "r" instead of "l" in the word "appealed" 
3) "u" instead  of "o" in word "without". 
All of these characters drawn mistakenly are similar to each other which makes it challenging for the generative models.

\begin{table}[hbt]
    \centering
    \resizebox{\linewidth}{!}{
    \begin{tabular}{|c|c|c|c|c|}
    \hline
         model  & FHD & ave. LD  & word accuracy\\
         \hline
         \hline
         WGAN &  874.76 &1.57&0.12\%\\
         \hline
         Pix2Pix &   814.24 &0.85&5.34\%\\
         \hline
         WGAN-GP &  68.57 &0.92&16.82\%\\
         \hline
         CTGAN &  51.55 &0.92& 15.48\%\\
         \hline
         SWGAN &  60.78 & 0.94&14.94\%\\
         \hline
         \hline
         Proposed method & 21.42  &0.36&55.36\%\\
         \hline
    \end{tabular}
    }
    \caption{Comparison of GAN models for IAM dataset}
    \label{tab:image2image_IAM}
\end{table}

\begin{figure*}[!t]
    \centering
    \includegraphics[width=\textwidth]{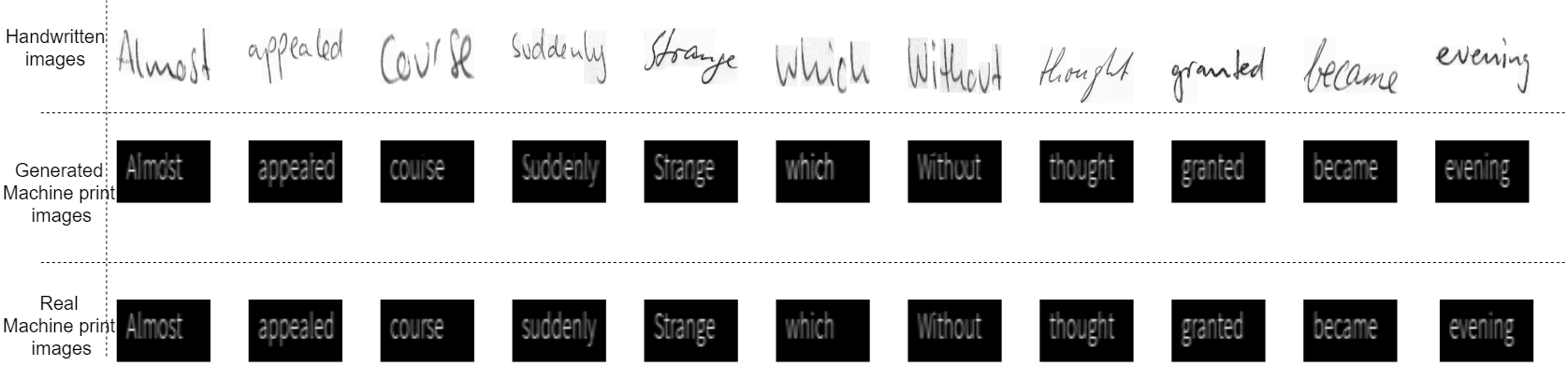}
    \caption{Some test examples of converting handwritten images to machine print ones. First row illustrate the handwritten images. Second row shows the generated machine print images and third shows the "real" machine print ones.}
    \label{fig:results}
\end{figure*}

\subsection{Effect of hidden dimension of LSTM on evaluation metrics}
This section talks about evaluating FHD, average LD and word accuracy metrics using different bidirectional LSTM's hidden dimensions in pretrained handwriting recognition models~\textsuperscript{\ref{note_1}}. It also shows that our model consistently outperforms baselines.
%
In Figure~\ref{fig:dim}, hidden dimension $\{16,32,64,128,256\}$ were used and results showed that 1) HW2MP-GAN, SWGAN, CTGAN and WGAN-GP models maintain consistency in their performance and 2) HW2MP-GAN was superior over all of them for all the hidden dimensions. 

\begin{figure*}[ht]
    \centering
  \begin{subfigure}[b]{0.67\columnwidth}
        \centering
        \includegraphics[width=\textwidth]{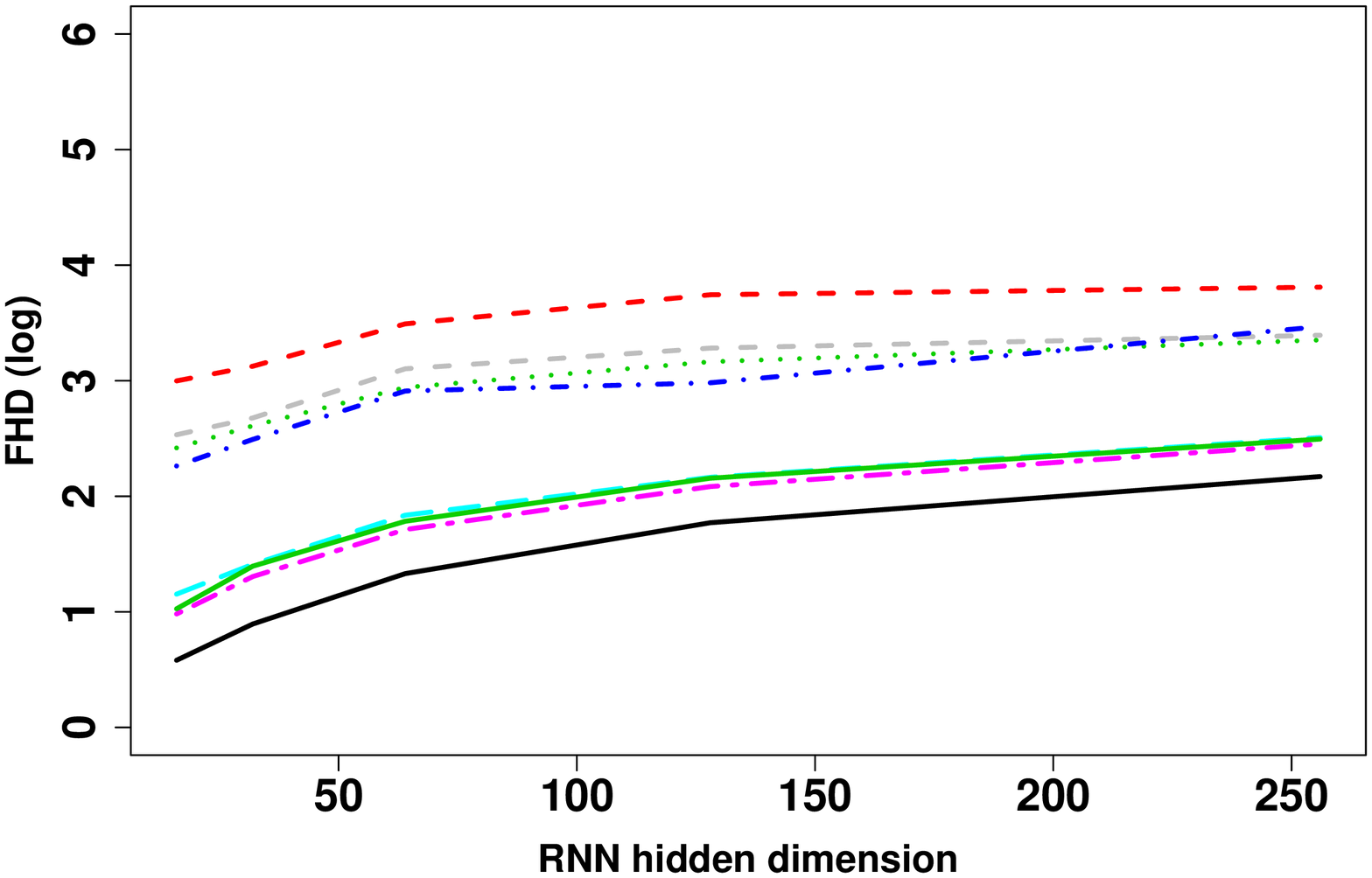}
        \label{fig:fid_rnn}
    \end{subfigure}
    \begin{subfigure}[b]{0.67\columnwidth}
        \centering
        \includegraphics[width=\textwidth]{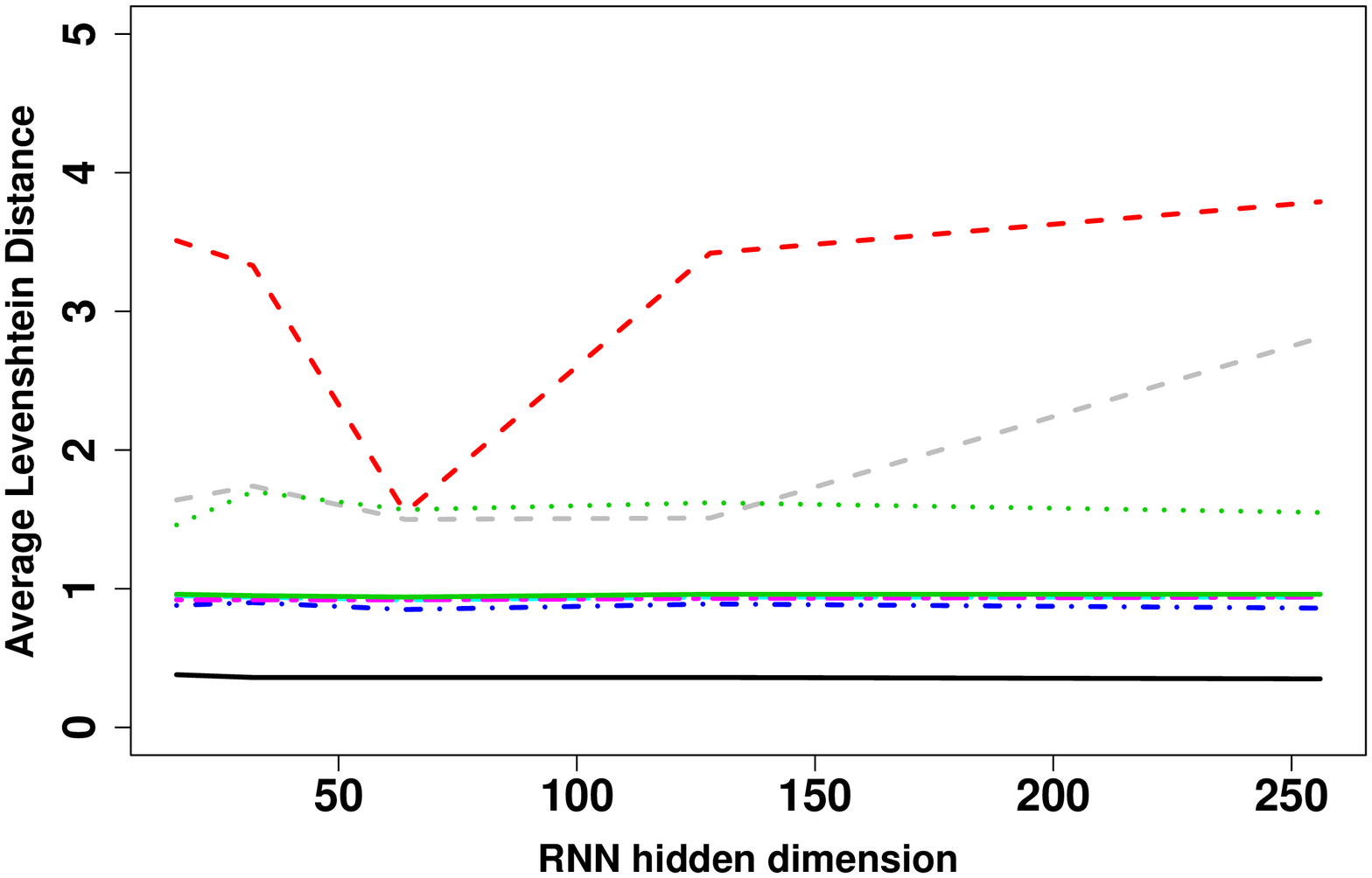}
        \label{fig:char_acc}
    \end{subfigure}
    \begin{subfigure}[b]{0.67\columnwidth}
        \centering
        \includegraphics[width=\textwidth]{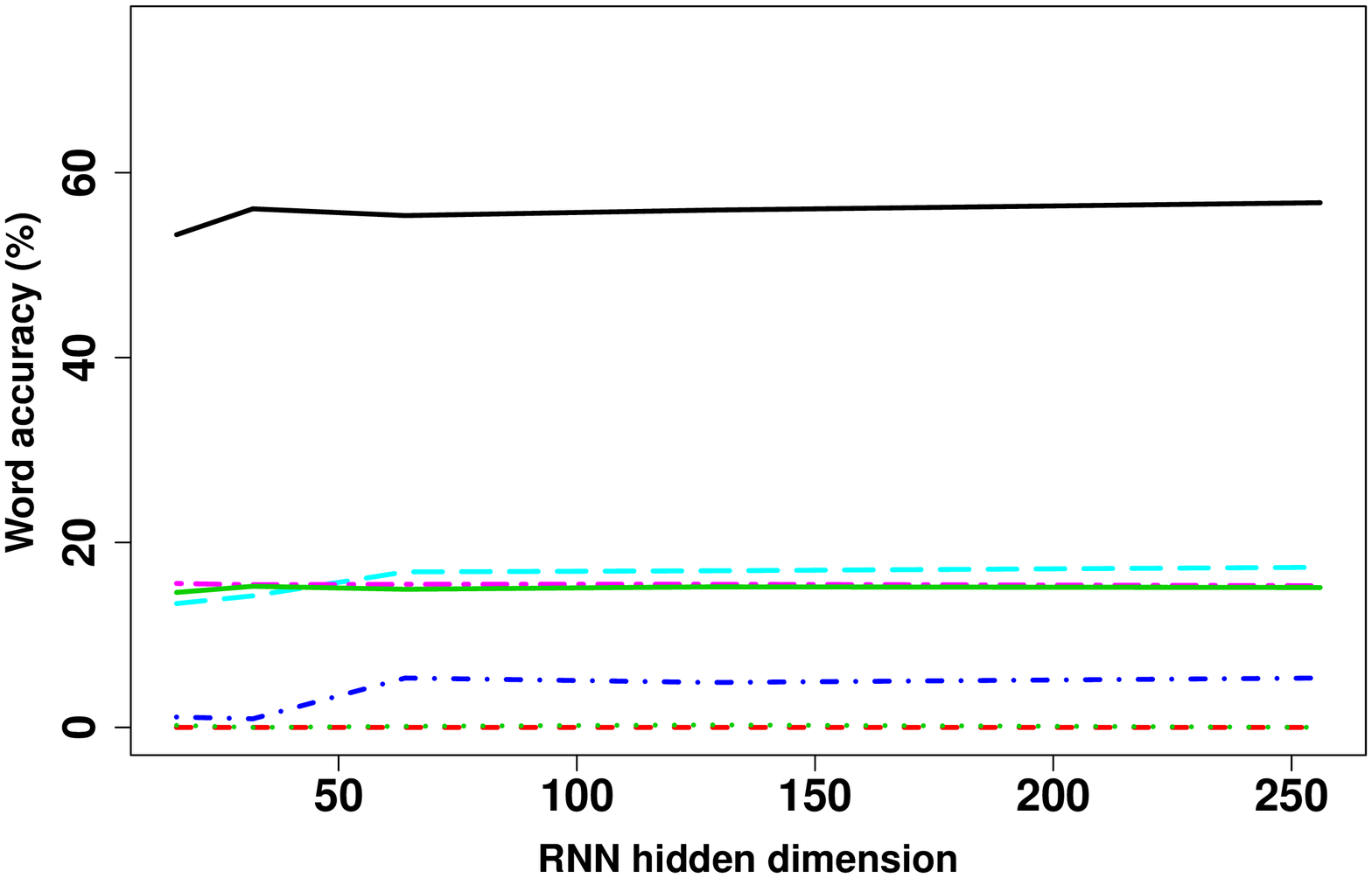}
        \label{fig:word_acc}
    \end{subfigure}\vspace{-0.5cm}
    \begin{subfigure}[b]{1.3\columnwidth}
        \centering
        \includegraphics[trim={0 17cm 0 2cm},clip,width=\textwidth]{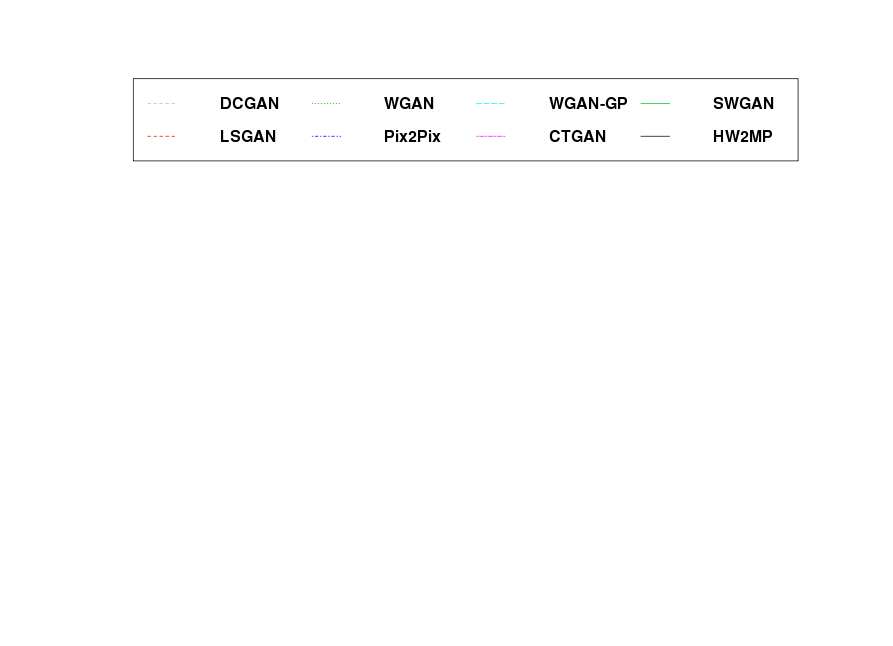}
        \label{fig:word_acc}
    \end{subfigure}
    \caption{Effect of hidden dimension of bidirectional LSTM in handwriting recognition on the performance of all GAN models in a) FHD b) Average Levenshtein distance c) Word accuracy }
    \label{fig:dim}
\end{figure*}


\subsection{Handwriting recognition reinforced with HW2MP-GAN}
We also evaluated the performance of the proposed attention-based handwriting recognition that has been discussed in Section~\ref{method:recognition} on the IAM dataset.  
The proposed model has been compared against these baselines: handwriting recognition (HWR) models trained by 1) handwritten images alone and 2) generated machine print only.
Table~\ref{tab:HWR_IAM} shows that the recognition model trained by handwritten text images gains a word accuracy of 84.08\% and 0.08 average LS, 62.12\% word accuracy and 0.3 average LD by only machine print.
Next, the proposed model trained using both results in 85.4\% word accuracy and 0.07 average LD. These results demonstrate the potential of exploiting the generated machine print images as an extra source of information to further boost the handwriting recognition task.

\begin{table}[hbt]
    \centering
    \resizebox{\linewidth}{!}{
    \begin{tabular}{|c|c|c|c|c|}
    \hline
         model & ave. LD  & word accuracy\\
         \hline
         \hline
         Handwritten images & 0.08 & 84.08\% \\
         \hline
          Generated machine print images only & 0.30 & 62.12\%\\
         \hline
         Generated machine print + handwritten images & \textbf{0.07} & \textbf{85.4\%}\\
         \hline
    \end{tabular}
    }
    \caption{Comparison of HWR models for IAM dataset}
    \label{tab:HWR_IAM}
\end{table}

\section{Conclusion}
\documentclass[main.tex]{subfiles}

In this paper, we have demonstrated the advantage of incorporating generative adversarial networks (GANs) in handwriting recognition problems. It has been shown that GAN-based document preprocessing such as handwritten to machine-print image transformation can further improve the accuracy of current handwritten recognition models. Our results on IAM database reveal the superiority of the proposed model on state-of-the-art conditional GAN models for handwritten image to machine-print image translation. Further improvements can be made over the proposed HW2MP-GAN model. Firstly, the model considers image preprocessing and handwritten recognition as separate tasks that can be combined into one. Secondly, current SWD with linear projections can be replaced by generalized SWD with nonlinear projections for more accurate estimate of distances between probabilities.




{\small
\bibliographystyle{ieee}
\bibliography{ref}

\begin{thebibliography}{10}\itemsep=-1pt

\bibitem{arjovsky2017principled}
M.~Arjovsky and L.~Bottou.
\newblock Towards principled methods for training generative adversarial
  networks, 2017.

\bibitem{arjovsky2017wasserstein}
M.~Arjovsky, S.~Chintala, and L.~Bottou.
\newblock Wasserstein generative adversarial networks.
\newblock In {\em International conference on machine learning}, pages
  214--223, 2017.

\bibitem{bahlmann2002online}
C.~Bahlmann, B.~Haasdonk, and H.~Burkhardt.
\newblock Online handwriting recognition with support vector machines-a kernel
  approach.
\newblock In {\em Proceedings Eighth International Workshop on Frontiers in
  Handwriting Recognition}, pages 49--54. IEEE, 2002.

\bibitem{blumenstein2002new}
M.~Blumenstein, C.~K. Cheng, and X.~Y. Liu.
\newblock New preprocessing techniques for handwritten word recognition.
\newblock In {\em Proceedings of the second IASTED international conference on
  visualization, imaging and image processing (VIIP 2002), ACTA Press,
  Calgary}, pages 480--484, 2002.

\bibitem{chen1993variable}
M.-Y. Chen, A.~Kundu, and S.~N. Srihari.
\newblock Variable duration hidden markov model and morphological segmentation
  for handwritten word recognition.
\newblock In {\em Proceedings of IEEE Conference on Computer Vision and Pattern
  Recognition}, pages 600--601. IEEE, 1993.

\bibitem{goodfellow2014generative}
I.~Goodfellow, J.~Pouget-Abadie, M.~Mirza, B.~Xu, D.~Warde-Farley, S.~Ozair,
  A.~Courville, and Y.~Bengio.
\newblock Generative adversarial nets.
\newblock In {\em Advances in neural information processing systems}, pages
  2672--2680, 2014.

\bibitem{graves2006connectionist}
A.~Graves, S.~Fern{\'a}ndez, F.~Gomez, and J.~Schmidhuber.
\newblock Connectionist temporal classification: labelling unsegmented sequence
  data with recurrent neural networks.
\newblock In {\em Proceedings of the 23rd international conference on Machine
  learning}, pages 369--376. ACM, 2006.

\bibitem{graves2007multi}
A.~Graves, S.~Fern{\'a}ndez, and J.~Schmidhuber.
\newblock Multi-dimensional recurrent neural networks.
\newblock In {\em International conference on artificial neural networks},
  pages 549--558. Springer, 2007.

\bibitem{graves2008novel}
A.~Graves, M.~Liwicki, S.~Fern{\'a}ndez, R.~Bertolami, H.~Bunke, and
  J.~Schmidhuber.
\newblock A novel connectionist system for unconstrained handwriting
  recognition.
\newblock {\em IEEE transactions on pattern analysis and machine intelligence},
  31(5):855--868, 2008.

\bibitem{graves2009offline}
A.~Graves and J.~Schmidhuber.
\newblock Offline handwriting recognition with multidimensional recurrent
  neural networks.
\newblock In {\em Advances in neural information processing systems}, pages
  545--552, 2009.

\bibitem{gulrajani2017improved}
I.~Gulrajani, F.~Ahmed, M.~Arjovsky, V.~Dumoulin, and A.~C. Courville.
\newblock Improved training of wasserstein gans.
\newblock In {\em Advances in neural information processing systems}, pages
  5767--5777, 2017.

\bibitem{isola2017image}
P.~Isola, J.-Y. Zhu, T.~Zhou, and A.~A. Efros.
\newblock Image-to-image translation with conditional adversarial networks.
\newblock In {\em Proceedings of the IEEE conference on computer vision and
  pattern recognition}, pages 1125--1134, 2017.

\bibitem{jo2019handwritten}
J.~Jo, H.~I. Koo, J.~W. Soh, and N.~I. Cho.
\newblock Handwritten text segmentation via end-to-end learning of
  convolutional neural network.
\newblock {\em arXiv preprint arXiv:1906.05229}, 2019.

\bibitem{karras2017progressive}
T.~Karras, T.~Aila, S.~Laine, and J.~Lehtinen.
\newblock Progressive growing of gans for improved quality, stability, and
  variation.
\newblock {\em arXiv preprint arXiv:1710.10196}, 2017.

\bibitem{kingma2014adam}
D.~P. Kingma and J.~Ba.
\newblock Adam: A method for stochastic optimization.
\newblock {\em arXiv preprint arXiv:1412.6980}, 2014.

\bibitem{kolouri2019generalized}
S.~Kolouri, K.~Nadjahi, U.~Simsekli, R.~Badeau, and G.~K. Rohde.
\newblock Generalized sliced wasserstein distances.
\newblock {\em arXiv preprint arXiv:1902.00434}, 2019.

\bibitem{kolouri2018sliced}
S.~Kolouri, P.~E. Pope, C.~E. Martin, and G.~K. Rohde.
\newblock Sliced-wasserstein autoencoder: an embarrassingly simple generative
  model.
\newblock {\em arXiv preprint arXiv:1804.01947}, 2018.

\bibitem{slicedkolouri2016}
S.~Kolouri, Y.~Zou, and G.~K. Rohde.
\newblock Sliced wasserstein kernels for probability distributions.
\newblock In {\em Proceedings of the IEEE Conference on Computer Vision and
  Pattern Recognition}, pages 5258--5267, 2016.

\bibitem{koshorek2018text}
O.~Koshorek, A.~Cohen, N.~Mor, M.~Rotman, and J.~Berant.
\newblock Text segmentation as a supervised learning task.
\newblock {\em arXiv preprint arXiv:1803.09337}, 2018.

\bibitem{kumar2013analytical}
G.~Kumar, P.~K. Bhatia, and I.~Banger.
\newblock Analytical review of preprocessing techniques for offline handwritten
  character recognition.
\newblock {\em International Journal of Advances in Engineering Sciences},
  3(3):14--22, 2013.

\bibitem{lecun1990handwritten}
Y.~LeCun, B.~E. Boser, J.~S. Denker, D.~Henderson, R.~E. Howard, W.~E. Hubbard,
  and L.~D. Jackel.
\newblock Handwritten digit recognition with a back-propagation network.
\newblock In {\em Advances in neural information processing systems}, pages
  396--404, 1990.

\bibitem{levenshtein1966binary}
V.~I. Levenshtein.
\newblock Binary codes capable of correcting deletions, insertions, and
  reversals.
\newblock In {\em Soviet physics doklady}, volume~10, pages 707--710, 1966.

\bibitem{mao2017least}
X.~Mao, Q.~Li, H.~Xie, R.~Y. Lau, Z.~Wang, and S.~Paul~Smolley.
\newblock Least squares generative adversarial networks.
\newblock In {\em Proceedings of the IEEE International Conference on Computer
  Vision}, pages 2794--2802, 2017.

\bibitem{marti2002iam}
U.-V. Marti and H.~Bunke.
\newblock The iam-database: an english sentence database for offline
  handwriting recognition.
\newblock {\em International Journal on Document Analysis and Recognition},
  5(1):39--46, 2002.

\bibitem{mirza2014conditional}
M.~Mirza and S.~Osindero.
\newblock Conditional generative adversarial nets.
\newblock {\em arXiv preprint arXiv:1411.1784}, 2014.

\bibitem{mori1999optical}
S.~Mori, H.~Nishida, and H.~Yamada.
\newblock {\em Optical character recognition}.
\newblock John Wiley \& Sons, Inc., 1999.

\bibitem{murdock2015}
M.~{Murdock}, S.~{Reid}, B.~{Hamilton}, and J.~{Reese}.
\newblock Icdar 2015 competition on text line detection in historical
  documents.
\newblock In {\em 2015 13th International Conference on Document Analysis and
  Recognition (ICDAR)}, pages 1171--1175, 2015.

\bibitem{o1993document}
L.~O'Gorman.
\newblock The document spectrum for page layout analysis.
\newblock {\em IEEE Transactions on Pattern Analysis and Machine Intelligence},
  15(11):1162--1173, 1993.

\bibitem{plamondon2000online}
R.~Plamondon and S.~N. Srihari.
\newblock Online and off-line handwriting recognition: a comprehensive survey.
\newblock {\em IEEE Transactions on pattern analysis and machine intelligence},
  22(1):63--84, 2000.

\bibitem{radford2015unsupervised}
A.~Radford, L.~Metz, and S.~Chintala.
\newblock Unsupervised representation learning with deep convolutional
  generative adversarial networks.
\newblock {\em arXiv preprint arXiv:1511.06434}, 2015.

\bibitem{reed2016generative}
S.~Reed, Z.~Akata, X.~Yan, L.~Logeswaran, B.~Schiele, and H.~Lee.
\newblock Generative adversarial text to image synthesis.
\newblock {\em arXiv preprint arXiv:1605.05396}, 2016.

\bibitem{ronneberger2015u}
O.~Ronneberger, P.~Fischer, and T.~Brox.
\newblock U-net: Convolutional networks for biomedical image segmentation.
\newblock In {\em International Conference on Medical image computing and
  computer-assisted intervention}, pages 234--241. Springer, 2015.

\bibitem{salimans2016improved}
T.~Salimans, I.~Goodfellow, W.~Zaremba, V.~Cheung, A.~Radford, and X.~Chen.
\newblock Improved techniques for training gans.
\newblock In {\em Advances in neural information processing systems}, pages
  2234--2242, 2016.

\bibitem{scheidl2018word}
H.~Scheidl, S.~Fiel, and R.~Sablatnig.
\newblock Word beam search: A connectionist temporal classification decoding
  algorithm.
\newblock In {\em 2018 16th International Conference on Frontiers in
  Handwriting Recognition (ICFHR)}, pages 253--258. IEEE, 2018.

\bibitem{shi2016end}
B.~Shi, X.~Bai, and C.~Yao.
\newblock An end-to-end trainable neural network for image-based sequence
  recognition and its application to scene text recognition.
\newblock {\em IEEE transactions on pattern analysis and machine intelligence},
  39(11):2298--2304, 2016.

\bibitem{shmelkov2018good}
K.~Shmelkov, C.~Schmid, and K.~Alahari.
\newblock How good is my gan?
\newblock In {\em Proceedings of the European Conference on Computer Vision
  (ECCV)}, pages 213--229, 2018.

\bibitem{starner1994line}
T.~Starner, J.~Makhoul, R.~Schwartz, and G.~Chou.
\newblock On-line cursive handwriting recognition using speech recognition
  methods.
\newblock In {\em Proceedings of ICASSP'94. IEEE International Conference on
  Acoustics, Speech and Signal Processing}, pages V--125. IEEE, 1994.

\bibitem{tappert1990state}
C.~C. Tappert, C.~Y. Suen, and T.~Wakahara.
\newblock The state of the art in online handwriting recognition.
\newblock {\em IEEE Transactions on pattern analysis and machine intelligence},
  12(8):787--808, 1990.

\bibitem{wei2018improving}
X.~Wei, B.~Gong, Z.~Liu, W.~Lu, and L.~Wang.
\newblock Improving the improved training of wasserstein gans: A consistency
  term and its dual effect.
\newblock {\em arXiv preprint arXiv:1803.01541}, 2018.

\bibitem{wu2019sliced}
J.~Wu, Z.~Huang, D.~Acharya, W.~Li, J.~Thoma, D.~P. Paudel, and L.~V. Gool.
\newblock Sliced wasserstein generative models.
\newblock In {\em Proceedings of the IEEE Conference on Computer Vision and
  Pattern Recognition}, pages 3713--3722, 2019.

\bibitem{xu1992methods}
L.~Xu, A.~Krzyzak, and C.~Y. Suen.
\newblock Methods of combining multiple classifiers and their applications to
  handwriting recognition.
\newblock {\em IEEE transactions on systems, man, and cybernetics},
  22(3):418--435, 1992.

\bibitem{zhang2007stroke}
Q.~Zhang, H.~A. Rowley, A.~A. Abdulkader, and A.~Guha.
\newblock Stroke segmentation for template-based cursive handwriting
  recognition, Nov.~27 2007.
\newblock US Patent 7,302,099.

\end{thebibliography}
}

\end{document}